\pdfoutput=1

\documentclass[11pt]{article}

\usepackage[]{acl}

\usepackage{times}
\usepackage{latexsym}

\usepackage[T1]{fontenc}

\usepackage[utf8]{inputenc}
\usepackage[font=small,labelfont=bf]{caption}
\usepackage{microtype}
\usepackage{array}
\usepackage{float}
\usepackage{graphicx}
\title{AIR-JPMC@SMM4H’22: Classifying Self-Reported Intimate Partner Violence in Tweets with Multiple BERT-based Models}

\author{Alec Candidato, Akshat Gupta, Xiaomo Liu, Sameena Shah\\
        J.P. Morgan AI Research \\
        \texttt{
            aleccandidato@gmail.com,
            \{akshat.x.gupta,
            } \\
        \texttt{
            xiaomo.liu, 
            sameena.shah\}@jpmchase.com
        }
}

\begin{document}
\maketitle
\begin{abstract}
    This paper presents our submission for the SMM4H 2022-Shared Task on the classification of self-reported intimate partner violence on Twitter (in English). The goal of this task was to accurately determine if the contents of a given tweet demonstrated someone reporting their own experience with intimate partner violence. The submitted system is an ensemble of five RoBERTa models each weighted by their respective F1-scores on the validation data-set. This system performed 13\% better than the baseline and was the best performing system overall for this shared task.
\end{abstract}

\section{Introduction}
    Intimate Partner Violence (IPV) refers to any form of physical or emotional abuse or aggression within a romantic relationship. Many people use social media, particularly twitter, as a means of self-reporting their experiences with IPV. Evidently, being able to parse through social media for these instances of self-reported IPV domestic violence is vital for finding victims most in need of resources and support. While scraping tweets for IPV-related content is relatively simple by keyword search, properly labeling and characterizing whether these tweets refer to self-reported IPV proves to be more challenging.
    
    It is already known that using BERT models or transformer-based models already produces state-of-the-art results for the field of Natural Language Understanding \citep{devlin-etal-2019-bert}. In fact, shared tasks from previous years have already demonstrated successful results incorporating this technology \citep{magge-etal-2021-overview}. This paper elaborates on our submission for the task of classifying self-reported intimate partner violence on Twitter (in English). The existing research surrounding the provided data-set also demonstrates the effectiveness of RoBERTa models \citep{https://doi.org/10.48550/arxiv.1907.11692}. Our final submission for this task is composed of a weighted ensemble of five RoBERTa-large models.
\section{Dataset}
    There were three different data-sets provided: training, validation, and test data-sets. The training and validation data-sets were labeled while the test data-set was not. All data-sets are composed entirely of tweets that are related to the topic of domestic violence. Among these tweets, 11\% are identified as self-reported IPV \citep{ALGARADI2022100217}. Table 1 shows the tweet distributions.
\begin{table}
    \centering
    \scalebox{0.8}{
        \begin{tabular}{c m{5em} m{5em} c}
        \hline
        \textbf{Split} & \textbf{Non-self-reported IPV} & \textbf{Self-reported IPV} & \textbf{Total}\\
        \hline
        Train & 4042 & 481 & 4523 \\
        Validation & 480 & 54 & 534 \\
        Test & --- & --- & 1291 \\
        \hline
        \end{tabular}
        \begin{tabular}{lc}
        \hline
        \end{tabular}
    }
    
    \caption{Frequency distribution for each dataset. The test dataset was unlabeled so the distribution is unknown.}
\end{table}

\section{Method}
    This submission utilizes an ensembling of multiple trained RoBERTa models which each used their best epoch by means of F1-score to outperform the previous results by almost a tenth of a point.
    The majority ensemble takes the mode model prediction of all five RoBERTa guesses for a given tweet. The weighted ensemble takes a precise linear combination of all five RoBERTa guesses based on each model's initial performance with the validation data-set. We have two defined classes: non-self-reported (0) and self-reported (1). Let us define ${P_e}(y = 1 \: | \: x)$ as the probability that the ensemble predicts class ${y} = 1$ for a given tweet, ${x}$. We can represent this explicitly as:
        \[  {P_e}(y = 1 \: | \: x) = \frac{\sum\limits_{i=1}^{n} a_i P_i(y = 1 \: | \: x)}{\sum\limits_{i=1}^{n} a_i}\]
    where $P_i$ represents an individual model's probability for a given tweet, $a_i$ represents the corresponding weight, and $n$ is the number of models being included in the ensemble. In the case of a majority vote, $a_i = 1$. Our submission uses $a_i = $ F1-score. Consequently, we can represent the probability for class 0 as ${P_e}(y = 0 \: | \: x) = 1 - {P_e}(y = 1 \: | \: x)$. The final prediction chooses the class with the highest probability.
    The task submission uses this weighted ensembling method because it can easily be combined with other models. Other non-transformer models were initially developed to being included with the ensemble but not ultimately used for the final submission.

\section{Preliminary Experiments}
    \begin{table}[h]
        \centering
        \scalebox{0.8}{
        
            \begin{tabular}{m{4em} m{4em} m{4em} m{4em} m{4em}}
            \hline
            \textbf{ } & \textbf{BERT-base} & \textbf{RoBERTa-base} & \textbf{BERT-large} & \textbf{RoBERTa-large} \\
            \hline
            Mean F1 & 0.718 & 0.749 & 0.710 & 0.779 \\
            Stdev & 0.013 & 0.017 & 0.014 & 0.013 \\
            \hline
            \end{tabular}
            \begin{tabular}{lc}
            \hline
            \hline
            \end{tabular}
        }
            
        \caption{Performance of BERT and RoBERTa classifiers on validation data. The F1 scores are averaged among all five models for each category. The standard deviation is also provided.}
        \label{tab:accents}
    \end{table}
    Five BERT-base \citep{devlin-etal-2019-bert} and five RoBERTa-base models \citep{https://doi.org/10.48550/arxiv.1907.11692} were first trained to test the initial effectiveness of transformer classifiers. We saved the best performing epoch for each model, based on a general accuracy metric. Then, five BERT-large and five RoBERTa-large models were trained. These models saved their best performing epoch in terms of F1-score instead. The F1-score for each model was determined based on their performance with the validation dataset. The results are shown in Table 2.
    
    The RoBERTa-large models performed significantly better than any other model we considered. Results from a majority voting ensemble and a weighted ensemble of five RoBERTa-large models were also calculated. They performed identically. Figure 1 shows the corresponding confusion matrix.
    \begin{figure}[h]
    \centering
    \includegraphics[scale=0.5]{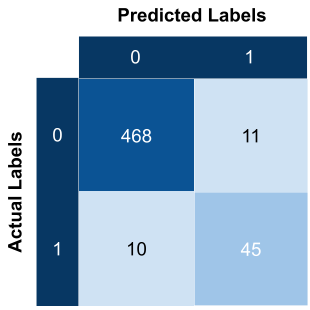}
    \caption{Confusion matrix for the RoBERTa-large ensembles on the validation data-set. This matrix represents the results for both the majority voting ensemble and weighted ensemble, since they performed identically to each other.}
\end{figure}

\section{Results}
    The performances of the designed ensemble compared to the median Codalab results and the best performing model from the original research are shown on Table 3. The RoBERTa ensemble performs significantly better than all other models by all metrics.
    \begin{table}
    \centering
    \scalebox{0.8}{
        \begin{tabular}{m{10 em} c c c}
        \hline
        \textbf{Classifier} & \textbf{F1-score} & \textbf{Precision} & \textbf{Recall}\\
        \hline
        Baseline \cite{ALGARADI2022100217} & 0.756 & 0.823 & 0.699 \\
        Median Task Performance & 0.763 & 0.790 & 0.716 \\
        \textbf{Our System} & \textbf{0.851} & \textbf{0.860} & \textbf{0.841} \\
        \hline
        \end{tabular}
        \begin{tabular}{lc}
        \hline
        \end{tabular}
    }
    \caption{Performance results for classifiers.}
    \end{table}
    The system that the median performer used is unknown at this time so no conclusions can be made comparing the results of the RoBERTa ensemble with that system. The best classifier from previous work was also built off of RoBERTa-large. Yet, this classifier achieved a 0.1 F1-score improvement. This may be because these RoBERTa models saved the epoch with the best F1-score rather than the best accuracy. Additionally, ensembling multiple transformer-based models has already proven to be effective \citep{jayanthi-gupta-2021-sj}.
\section {Conclusions}
    This study implements ensembling with transformer-based language models to drastically improve precision and recall for this task. This overall system outperforms previous metrics by nearly 13\%. This study also included initial pre-processing into part-of-speech frequency and the significance of narrative voice for self-reporting in tweets. Initial results suggests that tweets written in second or third person can help fine-tune against false positives. However, the final submission did not utilize any systems built off of this due to time limitations. Thus, these results are omitted from this paper. \citep{inproceedings} have already examined systems for narrative diegesis and point of view analysis. Ensembling transformer-based models with models trained off of narrative diegesis and point of view data may further improve results.
\bibliographystyle{acl_natbib}
\bibliography{custom}

\end{document}